\documentclass[letterpaper, 10 pt, conference]{ieeeconf}  

\IEEEoverridecommandlockouts                              

\title{\LARGE \bf
A Large Vision-Language Model based Environment Perception System for Visually Impaired People
}

\author{Zezhou Chen$^{1,2\dag}$, Zhaoxiang Liu$^{1,2\dag*}$, Kai Wang$^{1,2*}$, Kohou Wang$^{1,2}$, and Shiguo Lian$^{1,2*}$
\thanks{$^{1}$AI Innovation Center, China Unicom, Beijing 100013, China.}%
\thanks{$^{2}$Unicom Digital Technology, China Unicom, Beijing 100013, China. chenzz51, liuzx178, wangk115, wangzp103, liansg@chinaunicom.cn}%
\thanks{This work is supported by the Funding of Beijing Association of Science and Technology Outstanding Engineer Growth Plan}
\thanks{$\dag$Equal Contribution, $^{*}$Corresponding author}%
}

\usepackage{booktabs}
\usepackage{graphicx}
\usepackage{epstopdf}
\usepackage{multirow}
\usepackage{subfigure}
\usepackage{amsmath}
\usepackage{amsfonts}
\usepackage{color}
\usepackage{CJKutf8}
\usepackage{etoolbox}
\usepackage{tabularx}
\usepackage{float}
\makeatletter
\patchcmd{\@makecaption}
  {\scshape}
  {}
  {}
  {}
\makeatletter
\patchcmd{\@makecaption}
  {\\}
  {.\ }
  {}
  {}
\makeatother

\usepackage{array, caption, threeparttable}
\usepackage[font=small,labelfont=bf,labelsep=none]{caption}
\begin{document}
\begin{CJK}{UTF8}{gbsn}

\maketitle
\thispagestyle{empty}
\pagestyle{empty}

\begin{abstract}

  It is a challenging task for visually impaired people 
  to perceive their surrounding environment due to the
  complexity of the natural scenes. 
  Their personal and social activities are thus highly limited. 
  This paper introduces a Large Vision-Language Model(LVLM) 
  based environment perception system which helps them 
  to better understand the surrounding environment, 
  by capturing the current scene they face with a wearable device, 
  and then letting them retrieve the analysis results through the device. 
  The visually impaired people could acquire a global description of the scene 
  by long pressing the screen to activate the LVLM output, 
  retrieve the categories of the objects in the scene resulting 
  from a segmentation model by tapping or swiping the screen, 
  and get a detailed description of the objects 
  they are interested in by double-tapping the screen. 
  To help visually impaired people more accurately perceive the world, 
  this paper proposes incorporating the segmentation result of the RGB image as external knowledge
  into the input of LVLM to reduce the LVLM's hallucination. 
  Technical experiments on POPE, MME and LLaVA-QA90 show that the system could provide a more accurate description of the scene compared to Qwen-VL-Chat,  
  exploratory experiments show that the system helps visually impaired people to perceive the surrounding environment effectively. 

\end{abstract}

\section{INTRODUCTION}\label{sec:Intro}
Environment perception is essential for people's daily lives,
as it is a basic step for perceiving the surroundings and 
provides a reference for the next-step decision making~\cite{IT00}. 
While this seems easy for people with normal vision, 
it is quite challenging for visually impaired people who can hardly get a clear view of their surroundings~\cite{wang2017enabling}. 
According to the report of the World Health Organization, 
there have been 441.5 million people with severe vision impairment in the world, 
among which the majority are elderly~\cite{BR17} whose learning ability has degenerated~\cite{killeen2023objectively}. 
Without ease-of-use devices and supporting techniques, 
it is extremely difficult for them to get a comprehensive understanding 
of their surroundings as well as the things they are interested in. 
As a result, the quality of their social lives is thus greatly weakened.

While there have been some works proposed to assist 
the visually impaired people on completing single tasks, 
such as navigation~\cite{BL18}, reading~\cite{PT04}, 
object localization~\cite{lee2020hand, tseng2022vizwiz} and so on, 
few works have addressed the issue of giving them a comprehensive 
and precise description of the surroundings, 
while also accurately describing objects within the egocentric vision. 
The challenges lie in the following aspects, 
the environment is quite complicated with plenty of unorganized information. 
It is not easy to extract the information that is useful to users, 
and effectively organize it in a way that the global description of the surrounding environment 
as well as the local detailed description of the key components are preserved. 
Large Vision-Language Model(LVLM) is now flourishing in the research community, by exploiting powerful LVLM, 
users can get a description of a given image in any scene. 
However, the output descriptions of the LVLM are inconsistent with the input image, it is called hallucinations, 
and the LVLM claims non-existent objects and fails 
to describe the attribute of the object in the image accurately. 
Besides the hallucination problem, visually impaired people are also unable to accurately retrieve specific objects 
within a scene via only using LVLM. 

To address these challenges, this paper developed a system that is able to provide an immersive description of the scene to visually impaired people. 
The user can take a wearable device to capture the scene in front of him, then the analysis results are retrieved to the user 
through simple interactions over the screen of the device. Different from existing assistive egocentric vision methods, this paper proposed a unified multi-task framework  
that consists of a ViT~\cite{dosovitskiy2020image} based segmentation model 
and a LVLM, which could help visually impaired people get the global and local detailed description of the scene.
Users can obtain a scene description by long-pressing the device's screen, 
which activates the LVLM output. Tapping or swiping allows users to identify different objects within the scene 
based on the segmentation result, and detailed descriptions of specific objects can be accessed by double-tapping the screen. 
In light of the strong perception capability of the ViT-based segmentation model, inspired by~\cite{bi2019incorporating,peng2023check}, 
This paper proposed integrating the segmentation result of the RGB image as external knowledge into the input prompt of LVLM to reduce LVLM's hallucination 
to help visually impaired people more accurately perceive the world.


In summary, the contributions of this work are:
\begin{itemize}
  \item This paper proposes a unified multi-task framework that consists of a segmentation model and a LVLM, it could help visually impaired people get the global and local detailed description of the scene. 
  \item This paper proposes a training-free method that integrates the segmentation result as external knowledge into the input prompt of the LVLM to reduce the hallucination with relatively low computation cost compared to existing hallucination mitigation methods.
  \item The integration of these techniques into a mobile electronic system with which the user can retrieve global and local descriptions of the surrounding environment through simple interactions.
  \item This paper conducts technical experiments on mainstream benchmark datasets, results show that the method could effectively reduce the hallucination of the LVLM, and this paper conducts exploratory experiments on visually impaired people in various scenes, and results show that our system greatly improves users' experiences on perceiving the surrounding environment.
\end{itemize}

The rest of the paper is organized as follows: Section~\ref{sec:review} 
reviews the related work on systems for assisting 
visually impaired people's daily lives and methods for vision-based scene understanding. 
The proposed system is presented in Section~\ref{sec:sys}. 
Section~\ref{sec:layer} introduced the supporting algorithms for computing it. 
Experimental results are shown in Section~\ref{sec:results}, 
and conclusions are given in Section~\ref{sec:con}.
\section{Related work}\label{sec:review}

\subsection{Systems and interactions for visually impaired people's daily life tasks}

Plenty of technologies have emerged to assist visually impaired people with various tasks in their daily lives, 
ranging from the basic navigation~\cite{BL18, BL17}, collision detection~\cite{KC17,CK17}, 
reading~\cite{PT04} tasks to more complicated tasks like drawing~\cite{kurze1996tdraw}, 
playing~\cite{kane2011usable}, social contact~\cite{bennett2018teens} and so on. 
These technologies can be divided into two types: 
one is to enhance the users' ability to experience the world through other sensory organs, 
and the other is to help them perceive the world through sensors.

For the first kind of method, various devices and systems have been designed to enhance the user's haptic, auditory, taste and smell. 
For example, Chakraborty et al.~\cite{chakraborty2017flight} proposed a low-cost reading 
and writing system. Swaminathan et al.~\cite{swaminathan2016linespace} presented 
a sense-making platform for blind people by utilizing 3D printing technologies. 
Ducasse et al.~\cite{ducasse2016tangible} constructed tangible maps for visually impaired people. 
Bornschein et al.~\cite{bornschein2018comparing} proposed a computer-based drawing method 
for blind people with real-time tactile feedback. Holloway et al.~\cite{holloway2018accessible} 
presented accessible maps for blind people by comparing 3D printed models with tactile graphics. 
Brul{\'e} and Bailly~\cite{brule2018taking} took into account sensory knowledge to assist children with visual impairments learning geo-technologies. 
Kacorri et al.~\cite{kacorri2017people} trained personal object recognizers 
for people with visual impairments particularly. Leiva et al.~\cite{leiva2017synthesizing} 
synthesized stroke gestures of visually impaired people and enlarged the gestures database. 
Anyway, none of these works have considered the more complicated problem 
of simulating the visual ability to help visually impaired people understand the layout and properties of the surroundings.

For the second kind of method, information about the surroundings is conveyed to the user. 
The descriptions can be obtained manually through current computer vision technologies. 
Zhao et al.~\cite{zhao2018face} provided the users the information about identity, 
facial expressions and attributes of friends captured by their mobile phones. 
Advani et al.~\cite{AZ17} proposed to use of smart glasses and data glove 
to retrieve the product information in grocery shopping. 
While these tools are effective at perceiving human and grocery product information, 
their applications are still limited to restricted situations. 
The work~\cite{BL17} presented a system with multiple sensors mounted 
on wearable glasses to detect obstacles and inform the user via a beep sound during walking, 
and Bai et al.~\cite{BL18} further improved it by indicating the route to destination as well. 
Since these works are designed particularly for the navigation task, 
other surrounding information for comprehensive environment understanding cannot be provided.
Schoop et al.~\cite{ss18} utilized recent advances in real-time object detection 
and presented a wearable system that increases spatial awareness 
by detecting relevant objects in live 360-degree video. 
However, the content it could perceive is still limited to recognized dynamic objects, 
and it is still difficult to comprehensively understand the surrounding environment with that system.

\subsection{Vision-based Scene Understanding}

The most common techniques used for vision-based scene understanding include: 
object detection, image semantic segmentation, and image caption.

In general, there are two different approaches for object detection. 
One regards it as a regression or classification problem, 
and makes a fixed number of predictions on grid (one stage)~\cite{RD18, LA15}. 
The other leverages a proposal network to find objects 
and then uses a second network to fine-tune these proposals 
and output a final prediction (two stages)~\cite{GD14, RH17}. 

Semantic segmentation with deep neural network has achieved impressive progress in recent years. 
FCN~\cite{LS15}, adapts classifiers for dense prediction by replacing the last fully-connected layer with convolution layers. 
SegNet~\cite{BK15} presents an encoder-decoder architecture 
and reuses the pooling indices from the encoder to decrease parameters; 
DeepLab~\cite{CP18} proposes Atrous Spatial Pyramid Pooling (ASPP) for exploiting multi-scale information. 
SETR~\cite{zheng2021rethinking} and Swin Transformer~\cite{liu2021swin} adopts a Transformer~\cite{vaswani2017attention} as the encoder part of their
segmentation model without utilizing any convolution layers.
Besides these RGB-based methods, there are also some works~\cite{HM16, JZ18} proposed 
to make use of the depth map to enhance the precision by fusing it into the segmentation network.


Image caption describes the content of a given image through generating sentences~\cite{AZ17, VT15, LX17}.  
Attention mechanism was introduced by Xu et al.~\cite{xu2015show}, incorporating a context vector 
to decode the input at each time step, thereby enhancing 
the relevance between image regions and words. 
L Yang et al.~\cite{yang2017dense} decomposes image descriptions into multiple descriptions of image regions, 
combining the descriptions of each region to form the final output. 
Prompt-based learning paradigm is introduced in the vision-language field 
that various tasks are unified as the language generation conditioned on the multi-modal contexts, 
i.e., the image and the prefix words~\cite{cho2021unifying, tsimpoukelli2021multimodal, wang2021simvlm}.
Specifically, LLaVA~\cite{liu2024visual}, BLIP2~\cite{li2023blip}, Flamingo~\cite{alayrac2022flamingo} 
designed different schemes to connect the image and language representations. 

However, the output descriptions of the prompt-based learning model are inconsistent with the input image, it is called hallucinations, 
such as the information of non-existent objects and inaccurate attributes of objects. 
Gunjal et al.~\cite{b10} introduced MHalDetect which can be used to train and benchmark models for hallucination detection and prevention. 
Liuetal.~\cite{b11} addressed this issue by introducing the first large and diverse visual instruction tuning dataset, named Large-scale Robust Visual(LRV)-Instruction. 
Luet al.~\cite{Lu-Evaluation} developed an evaluation module that automatically creates fine-grained and diverse visual 
question-answering examples to assess the extent of agnosia in multimodal large language model(MLLM) comprehensively. 
The above training-based methods require significant computational resources and specialized data and are fairly time-consuming. 
Yin et al.~\cite{woodpecker} propose a training-free method, 
they utilized auxiliary model outputs to correct hallucinations in the post-process stage of MLLMs. 
But this method actually comprises three pre-trained large-scale models which are not only time-consuming, 
but one of them is also proprietary and needs extra several times inferences, making them uneconomical with slow inference processes. 

In order to acquire robust and consistent analysis results of the surrounding environment for visually impaired people, 
different from the method in~\cite{woodpecker}, 
our method focuses on the pre-processing stage of MLLMs, utilizing semantic segmentation results as external knowledge to construct the input prompts for the LVLM to reduce hallucination without extra inferences of the large language model, and is more efficient.
Furthermore, our method addresses the issue of perceiving global and local description 
about the surrounding environment for visually impaired people through a simple interaction logic and a wearable device. 

\section{The proposed system}\label{sec:sys}

\subsection{Architecture}

\begin{figure} [htbp]
  \centering
  \subfigure[]{
    \centering
    \label{fig:hardware:a} 
    \begin{minipage}[b]{0.2\textwidth}
      \centering
      \includegraphics[scale=0.08]{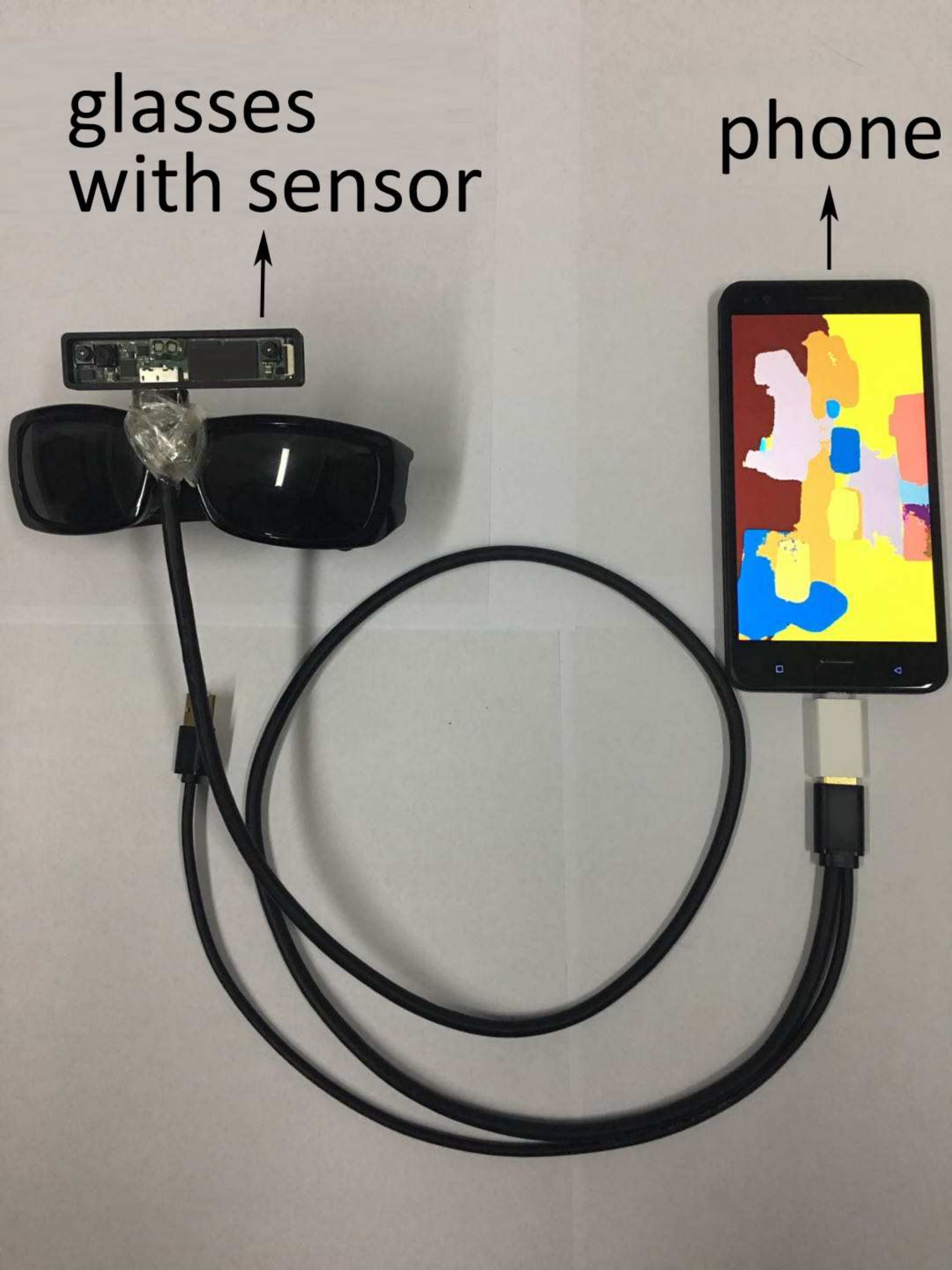}
    \end{minipage}}
  \subfigure[]{
    \centering
    \label{fig:hardware:b}
    \begin{minipage}[b]{0.2\textwidth}
      \centering
      \includegraphics[scale=0.08]{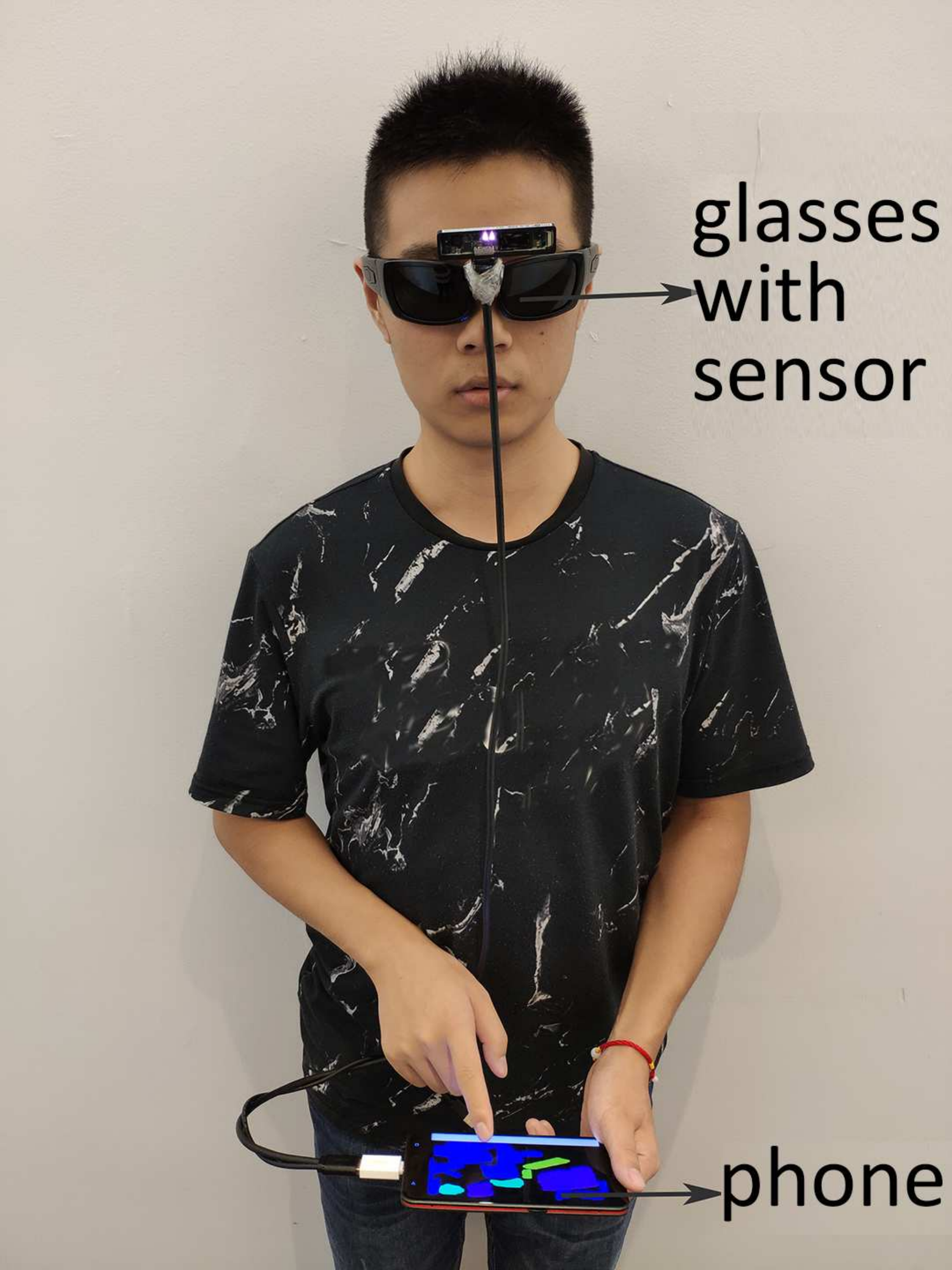}
    \end{minipage}}
  \caption{. (a) The terminal device of our system. (b) The user wearing the system. }
  \label{fig:hardware}
  \vspace{-5mm}
\end{figure}

The system is designed with a terminal-cloud architecture. 
The user can take a wearable device that connects to a smartphone 
to capture an image of the surrounding environment, as shown in Fig.~\ref{fig:hardware}. 
Unlike other systems~\cite{BL18,BL17} that perform basic neural network computations on a local CPU, 
this system sends images to the cloud for analysis, 
including a semantic segmentation model and a LVLM. 
This approach ensures reliable processing while keeping the wearable device lightweight and portable. 
Communication between the smartphone and the cloud is facilitated via the HTTP protocol. 
The design allows visually impaired people to interact easily 
with the system through gestures like long press, tap, swipe, 
and double-tap to retrieve and explore analysis results. 
Further details on the system's structure and computation process are covered in Section IV.

\begin{figure*}[htbp]
        \centering
        \hspace{-11mm}
        \includegraphics[scale=0.16]{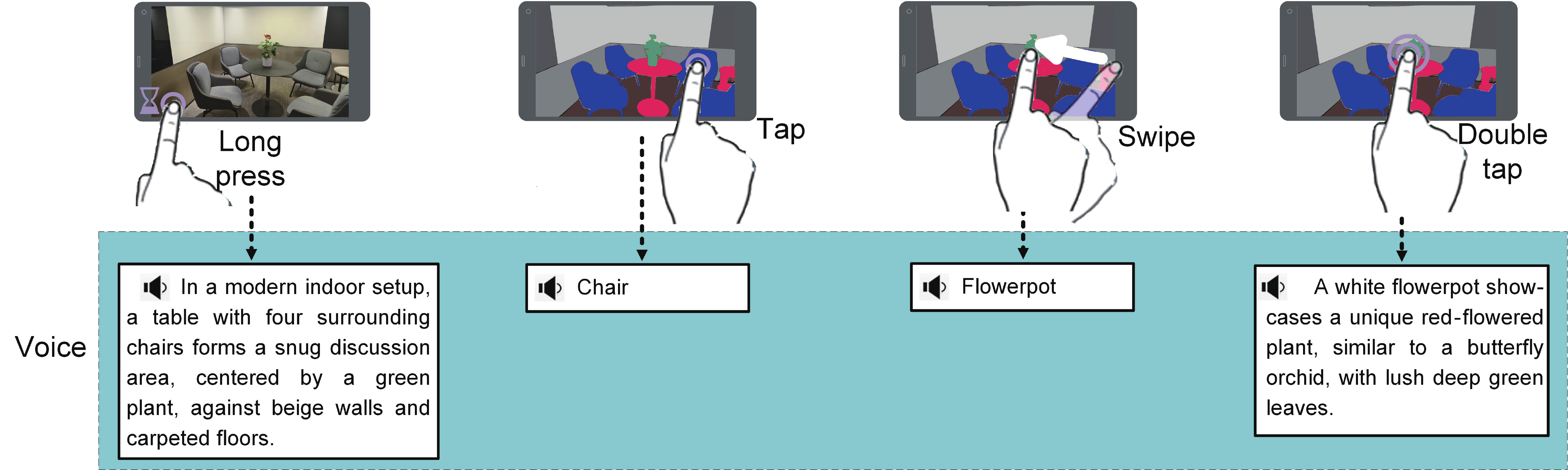}
        \caption{. The interaction procedure of the proposed system. The user long presses the screen first to take a picture of the current scene and the global description of the image will be played. After a long press action, the user could tap the chair area and the name of its class 'Chiar' is played. Then the user could swipe to the flowerpot and the name of its class 'Flowerpot' is played. Note that the volumes differ for objects with different distances. When the user double tapping the flowerpot area he detailed description of the flowerpot is played. }
        \label{fig:interaction}
        \vspace{-2mm}
\end{figure*}

\subsection{Hardware} 

For the scene-capturing device, the system used an RGB-D (Red, Green, Blue, and Depth) sensor bonded to the frame of a sunglass. The power of the sensor is supplied by a touch-screen smartphone via a USB cable, through which the data is transmitted at the same time. The phone provides data collection and communication, touch interaction and voice play functions, and serves as the main interaction device in the system. It should be mentioned that we use the speaker on the phone to provide the sound and voice feedback, instead of using an earphone. This is based on the observation that visually impaired people are reluctant to use earphones for the protection of their hearing, and using earphones to assist their understanding of the environment will also hinder them from getting aware of other signals in their daily lives.

\subsection{Interaction}

The interaction procedure of the system is shown in Fig.~\ref{fig:interaction}. 
A blend of touchscreen commands is used for interaction. 
Users begin by long-pressing on their mobile device to capture an image of the scene. 
The system then analyzes the image and notifies the user with a sound when information is ready, and the global description of the image is provided audibly. 
When the user taps or swipes the screen to explore different parts of the image, 
vibration feedback first notices the user that a new object has been touched, 
the name of the object is then provided through voice, with the voice volume indicating its distance. 
Double-tapping the screen at some location gives a detailed description of that specific object.
The system is designed for ease of use by visually impaired individuals, 
with voice feedback sped up to accommodate their enhanced auditory capability. 
Interaction continues as the user double-taps different areas to learn about other objects, 
with the option to start new sessions via a long press. 

While other interaction methods like voice control and air gestures were considered, 
they were not used due to outdoor noise and potential user fatigue, respectively. 

\section{The Proposed Scene Analysis Method}\label{sec:layer}

\subsection{Approach Overview}\label{sec:layer:class}

The scene analysis method of this paper employs a semantic segmentation model and a LVLM. 
The segmentation result is fully utilized to boost the capability of the LVLM.
Specifically, when visually impaired people long press the screen to get the global description of an image,
segmentation result of the RGB image is incorporated into the input prompt of the LVLM to reduce the LVLM's hallucination, 
helping visually impaired people more accurately perceive the world. 
When visually impaired people tap or swipe the screen to retrieve the category of a specific object, 
the segmentation result provides each object's category and area while the pixel-wise depth image is incorporated to adjust the voice volume. 
When visually impaired people perform double-tap action to get a detailed description of a specific object,
the segmentation result provides a priori knowledge about the object to guide the LVLM to output a detailed description of the object.

\subsection{Global Description Acquisition}\label{subsec:global}

\begin{figure} [htbp]
  \centering
  \hspace{0mm}
  \includegraphics[scale=0.14]{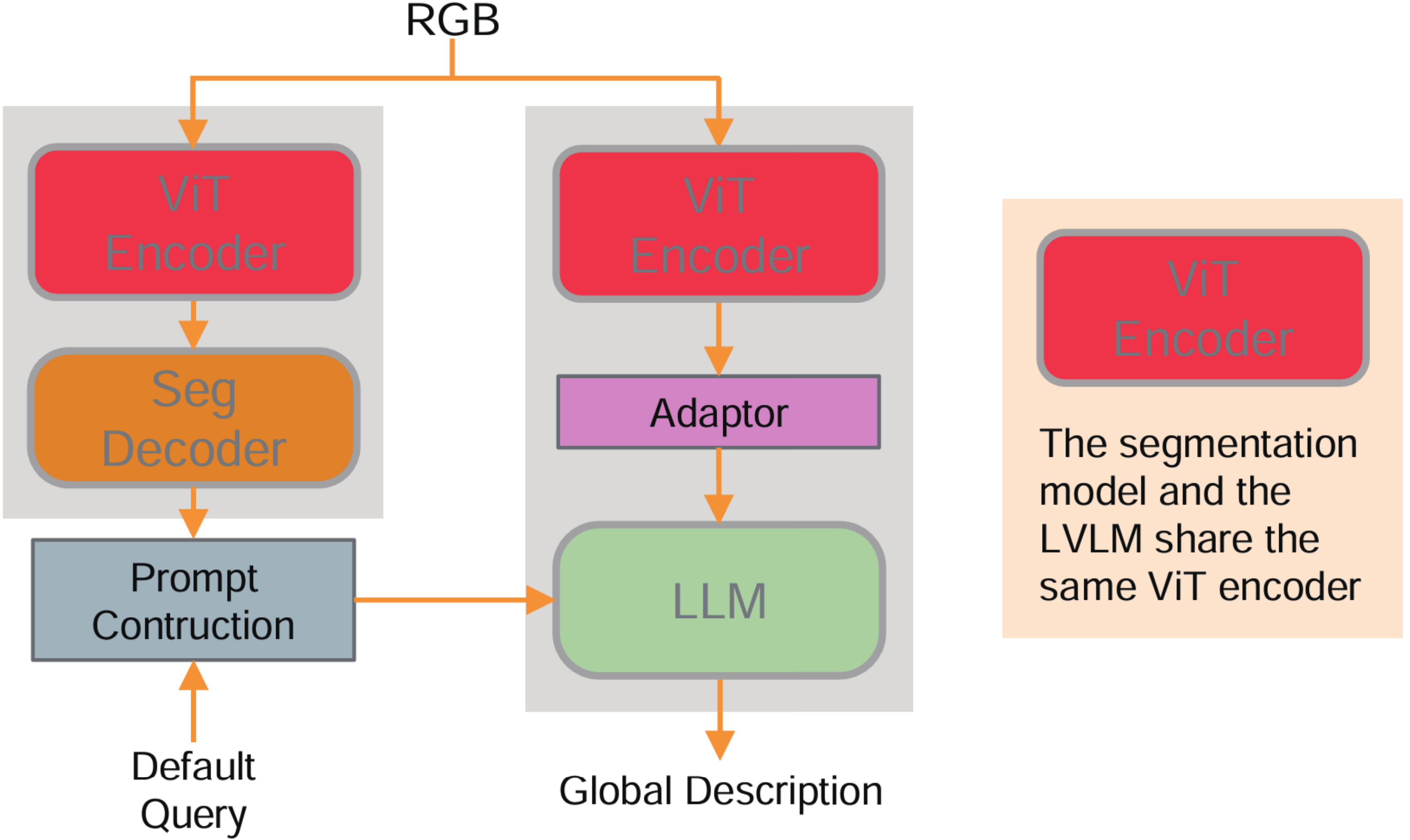}
  \caption{. The procedure of global description acquisition. }
  \label{fig:global_information}
\end{figure}

The procedure of global description acquisition is shown in Fig.~\ref{fig:global_information}.
Beginning with the user's long-press action, 
the RGB image $x\in\mathbb{R}^{H \times W \times 3}$ is fed into a ViT-based encoder 
to extract visual feature $x_v$. Subsequently, $x_v$ passes through a segmentation decoder, 
yielding a segmentation result.  

\begin{figure} [htbp]
  \centering
  \hspace{0mm}
  \includegraphics[scale=0.2]{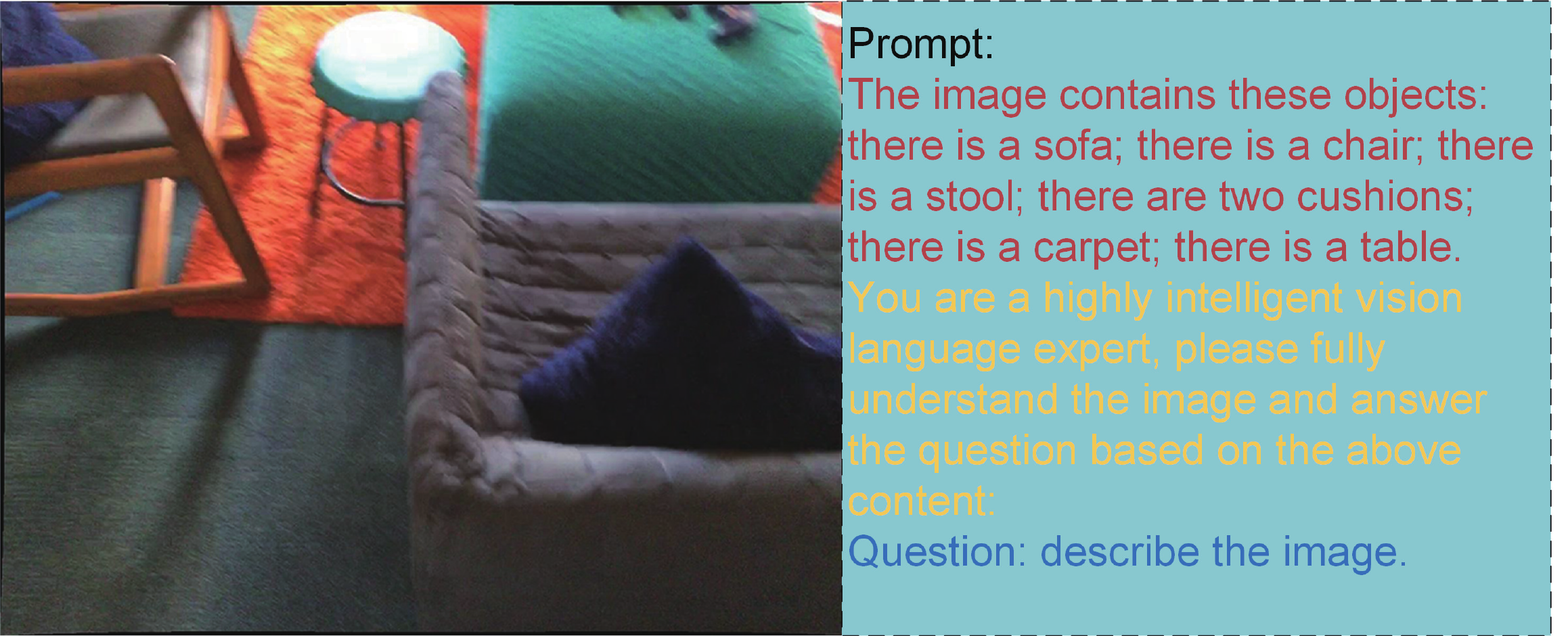}
  \caption{. Illustration of our prompt in the global description acquisition procedure. Red words are constructed based on the segmentation result, yellow words are the simple, ordinary prompt, and blue words are the default query. These segmented objects are integrated into one single sentence, which, alongside the default query, serves as the input of LLM. }
  \label{fig:global_prompt}
  \vspace{-3mm}
\end{figure}

In light of the strong perception capability of the segmentation model, 
we compile the input prompt based on the segmentation result of the image 
to enable the LVLM to describe the image precisely. 
Specifically, we traverse all the segmented objects and integrate them into a 
single sentence in the prompt. The detail is exemplified in Fig.~\ref{fig:global_prompt}.

Following this, an augmented prompt is constructed to guide the Large Language Model(LLM), 
and mitigate hallucinations in LVLM. After the prompt construction, 
the visual feature $x_v$ is aligned to the LLM via an adaptor, then concatenated with the augmented prompt, and finally fed into the LLM, 
resulting in an output that encompasses global description. 
It is worth noting that, both the LVLM and the segmentation model are computation intensive.
To reduce inference time, the segmentation model shares the same visual encoder with the LVLM, 
thus enabling the reuse of visual features during the inference.

\subsection{Object Category Retrieval}

When the user performs a tap or swipe action, the cloud server receives the position of the user's finger on the screen, 
and the category of the object corresponding to the position is acquired by retrieving the category information at the same position in the segmentation image, 
the area of the object is also retrieved. With the area of the object and pixel-wise depth image, 
the distance of the object can be calculated by averaging the depth values of all pixels within the area in the depth image, 
the average distance could be used to adjust the volume of the object's voice playback. 

\subsection{Local Description Acquisition}

\begin{figure} [htbp]
  \centering
  \hspace{4mm}
  \includegraphics[scale=0.15]{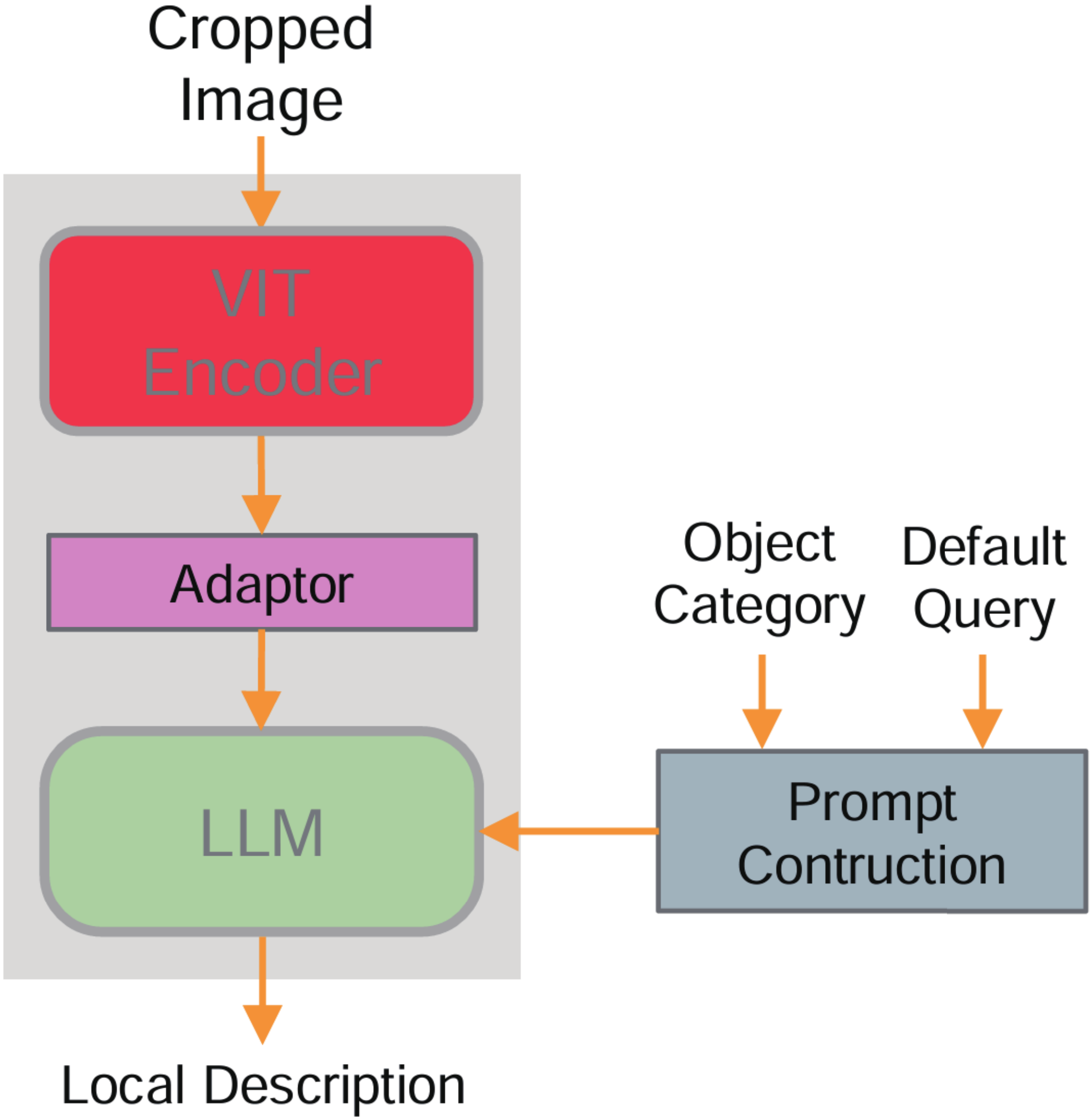}
  \caption{. The procedure of local description acquisition. }
  \label{fig:local_information}
  \vspace{-3mm}
\end{figure}

\begin{figure} [htbp]
  \centering
  \subfigure[]{
    \centering
    \label{fig:layerresultobj:a} 
    \begin{minipage}[b]{0.2\textwidth}
      \centering
      \includegraphics[scale=0.9]{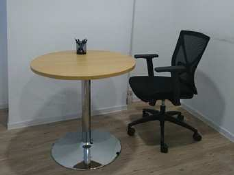}
    \end{minipage}}
  \subfigure[]{
    \centering
    \label{fig:layerresultobj:b}
    \begin{minipage}[b]{0.2\textwidth}
      \centering
      \includegraphics[scale=0.1078]{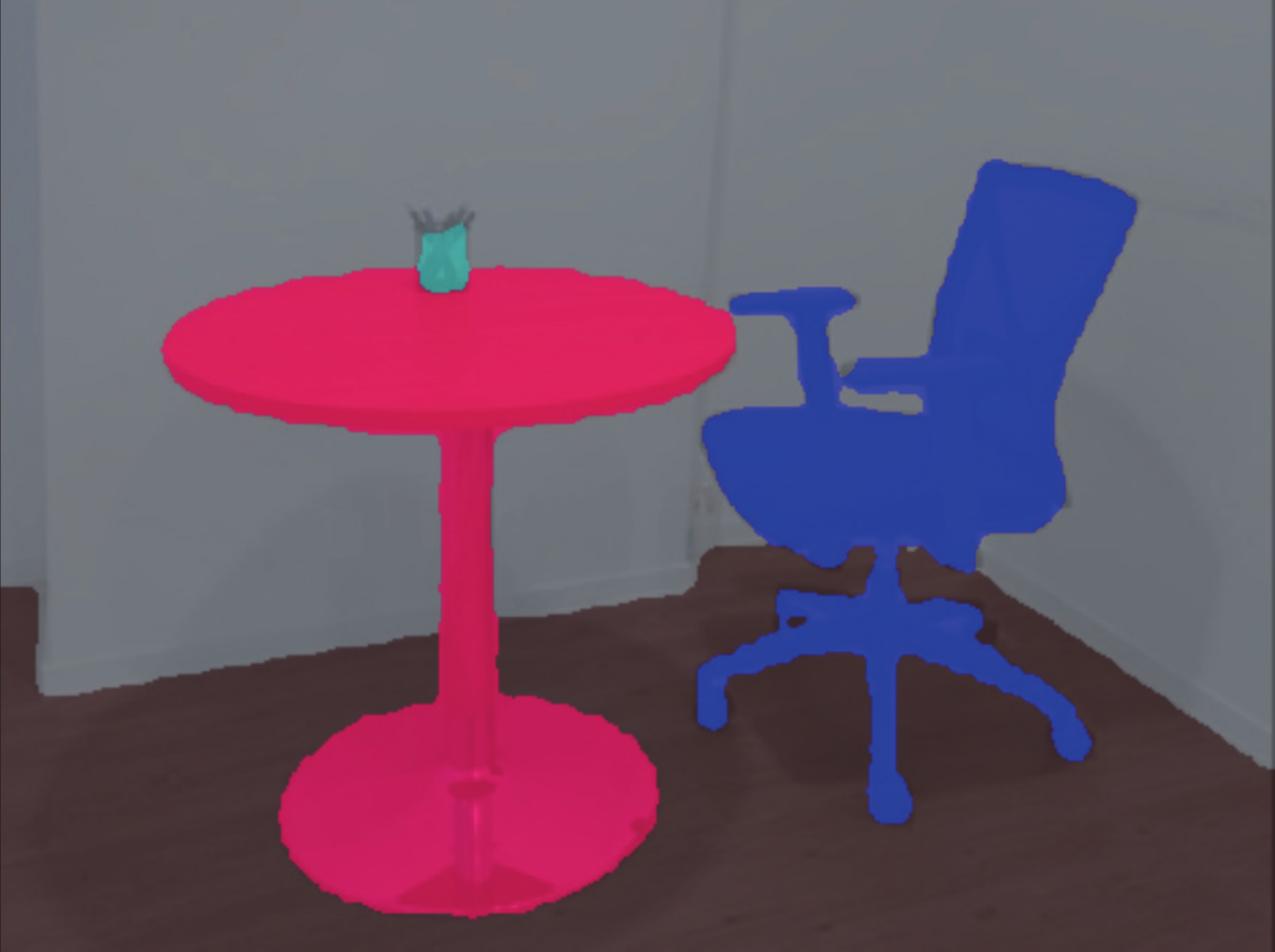}
    \end{minipage}}
  \subfigure[]{
    \centering
    \label{fig:layerresultobj:c}
    \begin{minipage}[b]{0.4\textwidth}
      \centering
      \includegraphics[scale=0.29]{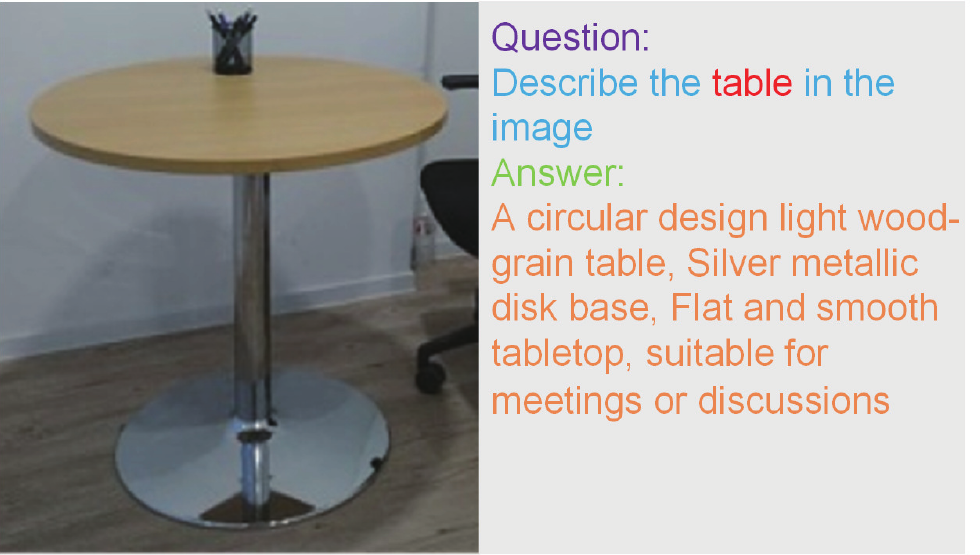}
    \end{minipage}}
  \caption{. An example of the segmentation and object analysis result. (a) The input image. (b) Segmentation result. (c) The table area is double-tapped and cropped, then a question including the "table" word is constructed, and finally the detailed description of the table is generated.}
  \label{fig:layerresultobj}
\end{figure}

The procedure of local description acquisition is shown in Fig.~\ref{fig:local_information}.
When the user performs a double-tap action, the bounding rectangle of the tapped object can be obtained based on the tap position on the screen and the segmentation result, 
then the bounding rectangle area of the object from the RGB image is cropped to construct 
the input image for the LVLM. And similar to the Section~\ref{subsec:global}, 
the cropped image is fed into the ViT-based encoder to extract visual feature 
and the visual feature is aligned to the LLM's embedding space via an adaptor, 
then concatenated with a prompt and fed into the LLM, 
resulting in an output that encompasses a detailed description of the object. 
An example of the segmentation and object analysis result is shown in Fig.~\ref{fig:layerresultobj}.

\section{Experiments}\label{sec:results}

\subsection{Implementation Detail}

For the LVLM, we choose the Qwen-VL-Chat~\cite{bai2023qwenvl} model which consists of a ViT-based encoder, an adaptor, and a LLM.
For the segmentation model, we choose SETR~\cite{zheng2021rethinking} which shares the same visual encoder with LVLM.  
We finetune the SETR's decoder on the VizWiz dataset~\cite{tseng2022vizwiz} while its encoder is frozen during the finetuning. 
The VizWiz dataset consists of annotations for images taken by visually impaired people
who authentically were trying to learn about their visual surroundings. 
 
We used a workstation with a 32-core CPU and an A100 GPU as the cloud server. 
And we calculated the average response time for each interaction action.
The average response time of the long press action is around 4400 ms, 
while for the tap/swipe action, it is 400 ms, and for the double-tap action is around 2800 ms. 
Although we did not achieve real-time feedback for the participants, 
the interaction process remains fluent enough. 

In order to evaluate the utility of
the system in various scenes for the visual impairments community, 
twelve visually impaired people were invited as participants, 
including eight males and four females. Among them, four are totally blind, 
and the other eight are with visual impairments in different degrees; 
seven are early blind, and the remaining five are late blind. 
Eight of them are with the age older than 50. 
All of the participants are familiar with smartphone operations. 
We have obtained approval from institutions for visually impaired people and all visually impaired participants for the experiments. 

These experiments include objective and subjective components. 
The objective experiments are conducted on mainstream benchmark datasets including
POPE~\cite{li2023evaluating}, MME~\cite{MME} and LLaVA-QA90~\cite{liu2024visual} to validate the 
effectiveness and superiority of our method.
The subjective experiments are evaluated by the participants in four scenes: 
an office, a shopping mall, a street and a park, 
all of which are the typical scenes that they work or live in. 
The participants are free to walk around and perform the tasks as they wish, 
and an observer with normal vision is arranged to accompany each user 
and record the test result and user's feedback. 
An evaluation for a participant in each scene is set to around 20 minutes. 
If the interval between two operations on the device is longer than 30s, 
the participant would be reminded by the observer.

\subsection{Technical Evaluation}

\noindent\textbf{POPE.} The POPE~\cite{li2023evaluating} initiative aims to gauge the tendency of MLLMs to produce inaccuracies. 
It utilizes three varied sampling strategies: random, popular, and adversarial—to construct non-existent object samples. 
Random sampling draws from a pool of items not depicted in the image, 
while popular sampling selects frequently seen items not present. 
Adversarial sampling identifies items often found together but missing from the image. 
We sampled 500 images for each strategy, with each image corresponding to six questions-answers. 
We collected 1500 images and 9000 question-answers in total.

  \renewcommand{\arraystretch}{1.5}
  \begin{table}[htbp]
    \centering
    \captionof{table}{: Results on POPE using Qwen-VL-Chat as baseline model. w/Ours denotes LVLM responses generated by the proposed method. The best performances within each setting are bolded. Our method achieves a universal advantage across the board.}
     \begin{tabular}{l l l l l l}
         \toprule
         setting & model & Accuracy&  Precision & Recall&  F1-Score  \\
         \midrule
         \multirow{2}{*}{adversarial} & Qwen & 0.842 & 0.897 & 0.772 & 0.830 \\
         \cline{2-6}
         & Ours & \textbf{0.862 } &\textbf{0.923 }&  \textbf{0.790} & \textbf{0.851 } \\
         \hline
         \multirow{2}{*}{random} & Qwen & 0.879 & 0.981 & 0.773 & 0.864 \\ 
         \cline{2-6}
         & Ours & \textbf{0.888 } &\textbf{0.982}&  \textbf{0.790} & \textbf{0.876} \\
         \hline
         \multirow{2}{*}{popular} & Qwen & 0.866 & 0.949 & 0.773 & 0.852 \\ 
         \cline{2-6}
         & Ours & \textbf{0.879 } &\textbf{0.960 }&  \textbf{0.791} & \textbf{0.867 } \\
         \bottomrule
     \end{tabular}
    \label{table_pope}
    \vspace{-3mm}
  \end{table}

For evaluation, we tested all these 9000 images. 
The questions balance between positive and negative samples at a 50-50 split. 
This approach casts object annotations as binary questions, 
centering on the evaluation of object hallucinations, 
with a particular emphasis on the aspect of existence. The selected MLLMs
will answer like "Is there a wine glass in the image?", and the answer will be
measured in a metric of Accuracy, Precision, Recall and F1 Score. 
The results are shown in Table~\ref{table_pope}. 

As can be seen from the results, the proposed method achieves 
an across-the-board performance improvement on all test sets and in all aspects. 
In detail, in all the asversarial, random, and popular testset, 
our method outperforms the baseline in all the accuracy, precision, recall, and f1-score, 

\renewcommand{\arraystretch}{1.5}
  \begin{table}[htbp]
    \centering
    \captionof{table}{: Results on using Qwen-VL-Chat as baseline model. w/Ours denotes LVLM responses generated by our method. The best performances within each setting are bolded. Our method achieves a universal advantage across the board.}
     \begin{tabular}{c c c }
         \toprule
         & Existence & Count \\
         \midrule
         Qwen & 185 & 140\\ 
         \hline
         Ours & \textbf{190} & \textbf{173}\\ 
         \bottomrule
     \end{tabular}
    \label{table_mme}
  \end{table}

\noindent\textbf{MME.} The MME~\cite{MME} is a comprehensive evaluation benchmark for MLLMs. 
The annotations of instruction-answer pairs are all manually designed 
to avoid data leakage. The concise instruction design can fairly compare MLLMs, 
with such an instruction, quantitative statistics can also be easily carried out. 
In this paper, considering the biggest concern of visually impaired people is perceiving the existence and count of objects in the scene, so we use the object's existence and count examinations of the MME to evaluate the perception and cognition abilities of our method. 
We use 420 images with 840 instruction-answer pairs in total, 
and show the results in Table~\ref{table_mme}. 
As can be seen from the table, our proposed method achieves an across-the-board performance 
improvement. 

\renewcommand{\arraystretch}{1.5}
  \begin{table}[htbp]
    \centering
    \captionof{table}{: Results on using Qwen-VL-Chat as baseline model. w/Ours denotes LVLM responses generated by our method. The best performances within each setting are bolded. Our method achieves a universal advantage across the board.}
     \begin{tabular}{c c c}
         \toprule
         & average accuracy & average detailedness\\
         \midrule
         Qwen & 6.38 & 5.25 \\
         \hline
         Ours & \textbf{7.25} & \textbf{6.25} \\ 
         \bottomrule
     \end{tabular}
    \label{table_example}
  \end{table}

\noindent\textbf{LLaVA-QA90.} The dataset contains randomly selected 30 images for COCO-Val-2014, 
and for each image, three types of questions (conversation, detailed description, complex reasoning) 
are generated using the proposed data generation pipeline in~\cite{liu2024visual}. 
Specifically, we sample 13 description-type queries that are paraphrased in various forms 
to instruct an MLLM to describe an image. The results can be seen from the Table~\ref{table_example}. 
The results show that the proposed method has also achieved superior performance in this dataset.

The above experiments' results show that the segmentation model which specializes in identifying, localizing and segmenting objects, 
could provide more accurate information about the existence and the number of objects, 
and this information could serve as external knowledge to guide the LVLM to generate a more accurate description of the scene and reduce its hallucinations.

\subsection{Exploratory Evaluation}

Questions of the survey include open-ended answers and 5-point Likert scales. 
Likert scale questions were phrased as follows: 
\emph{
Q1: The overall perception of the surrounding environment is useful. 
Q2: The information of specific objects is useful for users.
Q3: The information is computed and transmitted quickly.
Q4: The interaction of the system is easy to use.  } 
The first three questions were answered by the visually impaired people in different scenes, 
and the fourth question was answered after they completed the experiment in four scenes.

\begin{figure} [htbp]
  \centering
  \subfigure[]{
    \centering
    \label{fig:ques:a} 
    \begin{minipage}[b]{0.23\textwidth}
      \centering
      \includegraphics[scale=0.09]{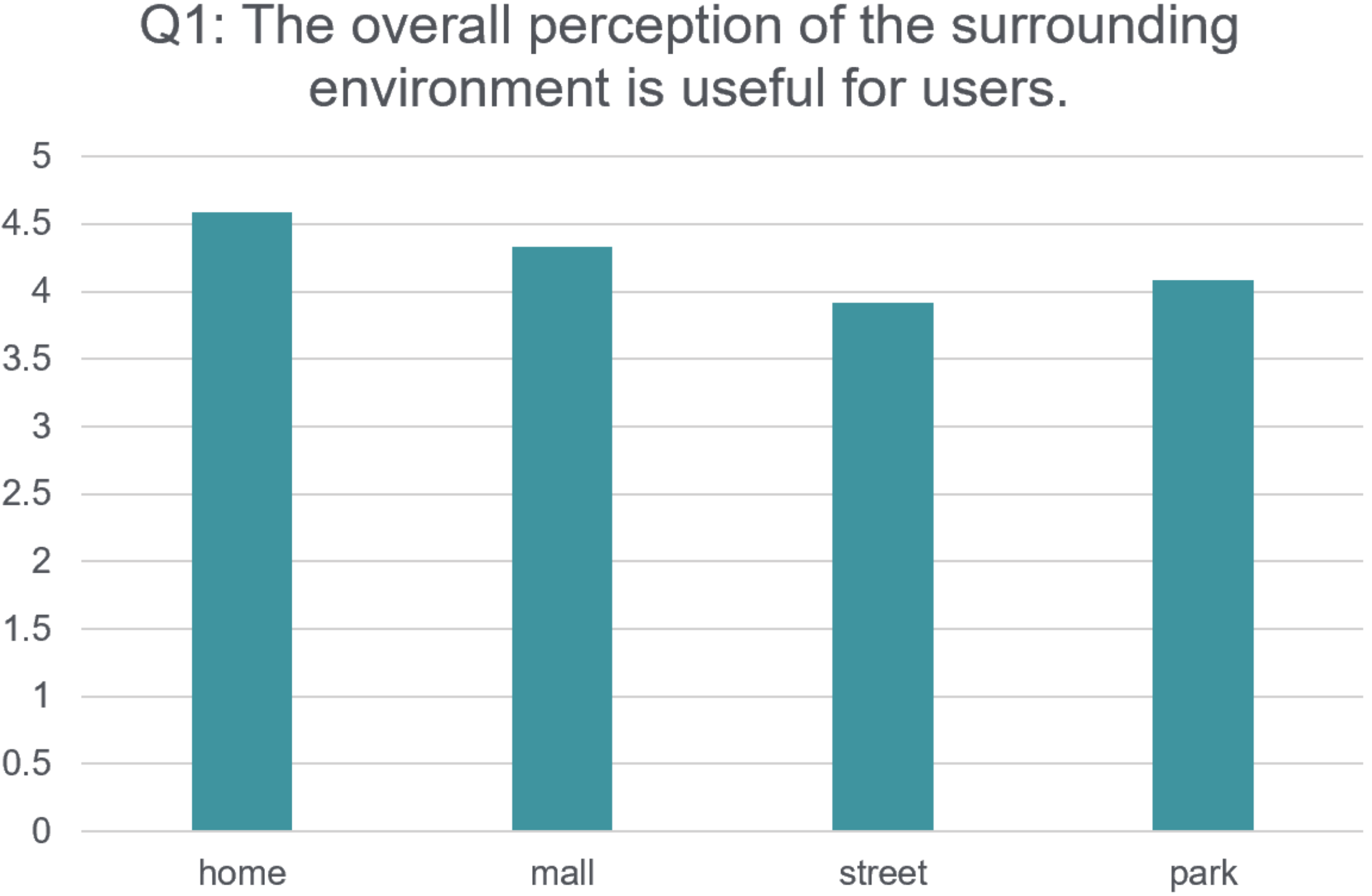}
    \end{minipage}}
  \subfigure[]{
    \centering
    \label{fig:ques:b}
    \begin{minipage}[b]{0.23\textwidth}
      \centering
      \includegraphics[scale=0.09]{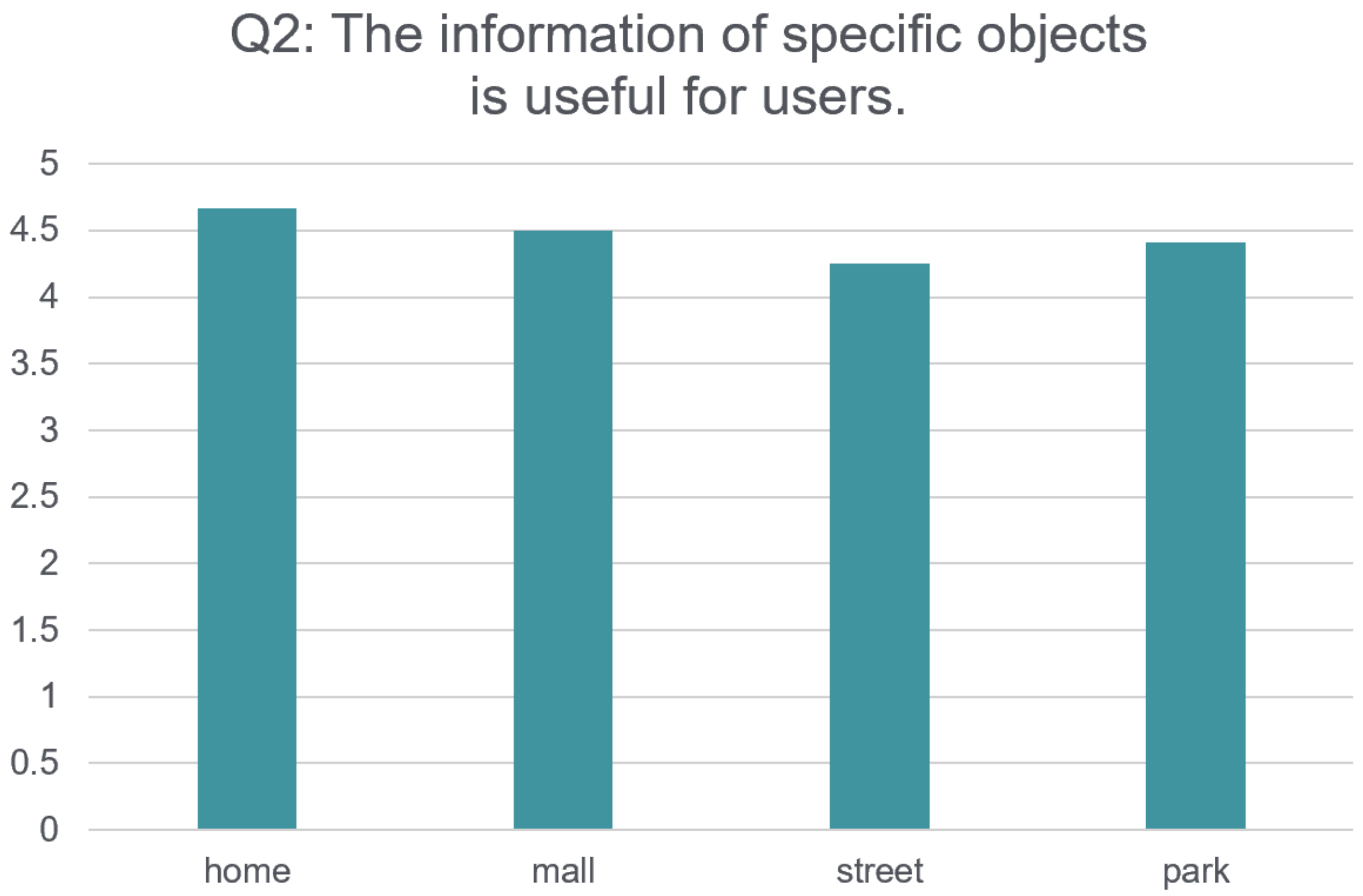}
    \end{minipage}}
  \subfigure[]{
    \centering
    \label{fig:ques:c}
    \begin{minipage}[b]{0.23\textwidth}
      \centering
      \includegraphics[scale=0.09]{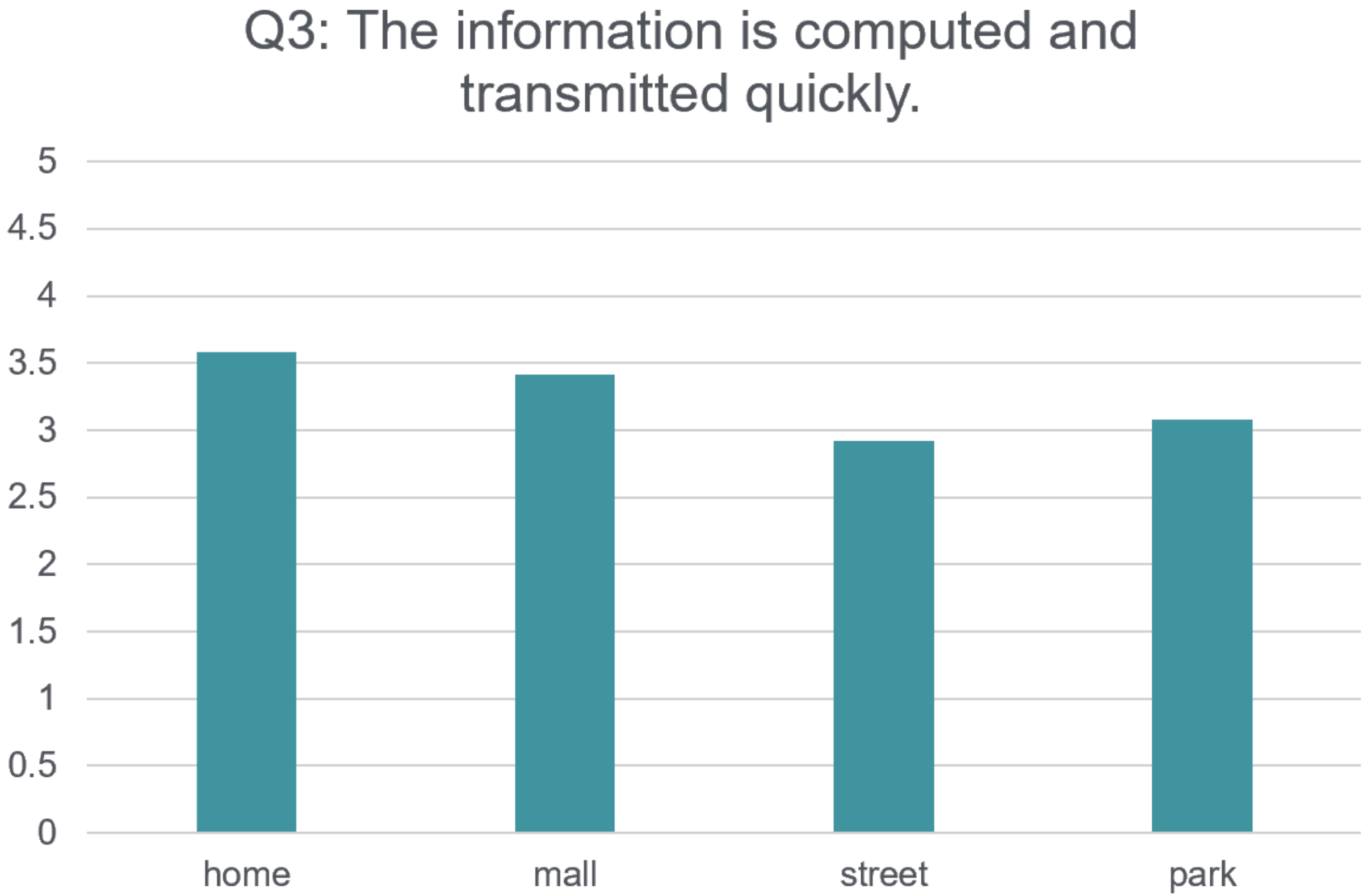}
    \end{minipage}}
  \subfigure[]{
    \centering
    \label{fig:ques:d}
    \begin{minipage}[b]{0.23\textwidth}
      \centering
      \includegraphics[scale=0.09]{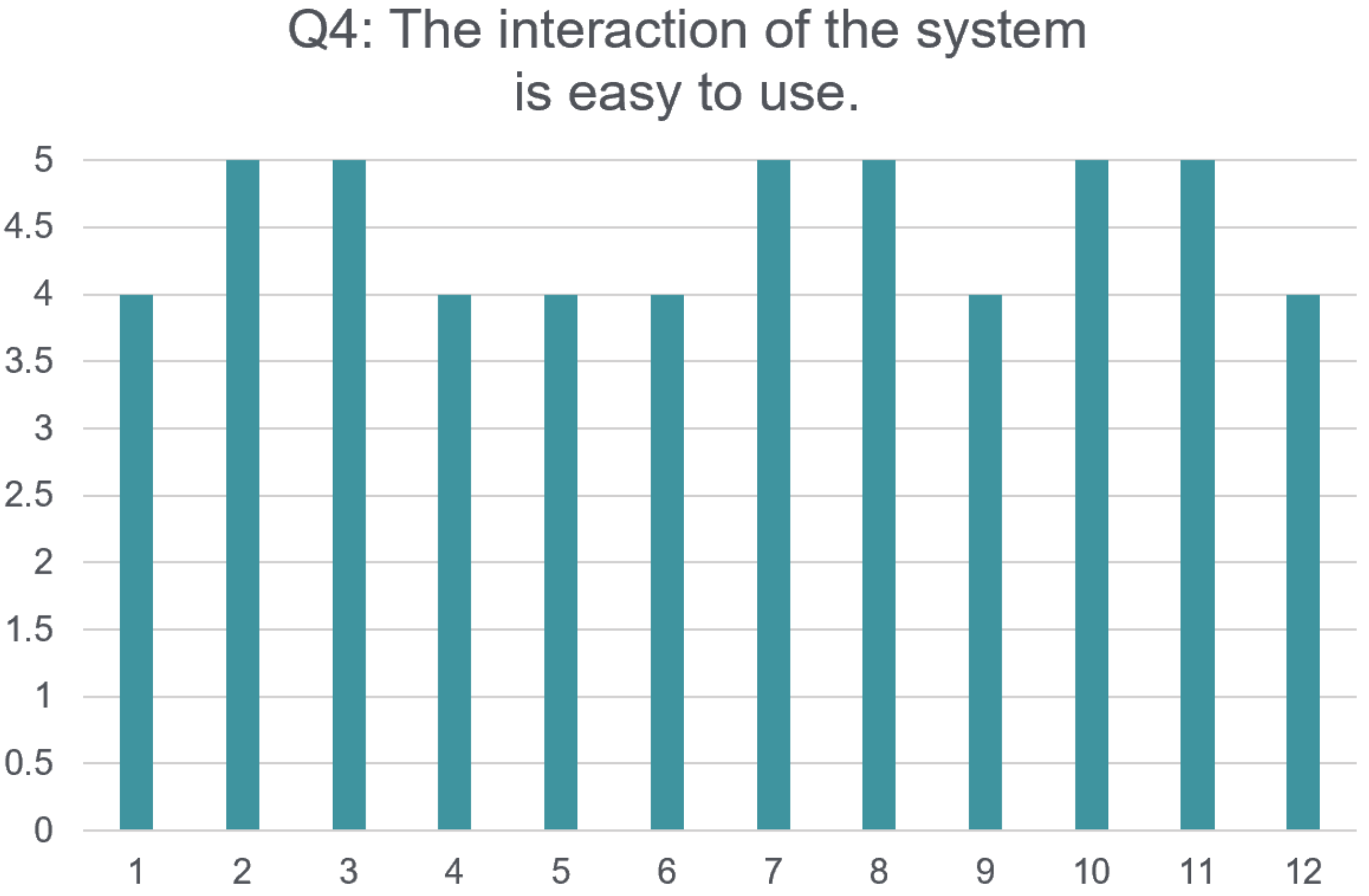}
    \end{minipage}}
  \caption{. Statistics of the perception evaluation and the interaction evaluation of using our system in different scenes. For Q1-Q3, each score represents the average score of the 12 participants in the corresponding scene. For Q4, each score represents the rating of the corresponding numbered participant. }
  \label{fig:ques}
\end{figure}

As shown in the results in Fig.~\ref{fig:ques}, 
participants generally expressed positive reactions towards using our system in the four scenes.

The high scores of \emph{Q1} show 
that the overall perception of the surrounding environment is useful. 
For example, in a room with chairs and a television,
the chairs and television, along with their approximate locations, 
are successfully perceived.  
In the street scene which contains cars and roads,  
the cars' approximate positions, as well as the direction and condition of the roads are analyzed.
The information is helpful for visually impaired individuals 
to understand the environment nearby like other people.

As for \emph{Q2}, participants felt they had perceived detailed attributes of objects.
For instance, when the object is a hat, 
our system not only provides basic information about the hat, such as color, 
but LVLM also attempts to offer additional descriptions including the purpose of the hat, 
the scene it is suitable for, whether it appears to be expensive, and so on. 

With regard to \emph{Q3}, participants said 
there may be instances where long processing time 
is incurred. This is because some analysis results are lengthy, 
and there are some places where the communication signals are not ideal. 
These problems will be alleviated with the progress of the hardware computing power.

The high scores of \emph{Q4} show that the interaction is user-friendly, 
with participants' feedback indicating 
that they quickly became accustomed to each action within the interaction logic.

From the statistics of feedback, 
we did not find explicit distinct between male 
and female on using our system. 
Young participants generally mastered 
the operations faster and explored broader spaces during the test. 
Although the older users are slow at learning the operations, 
they get used to it and gradually expand their test fields 
within a limited time.

\begin{figure} [htbp]
  \centering
  \subfigure[]{
    \centering
    \label{fig:userstudycase:a} 
    \begin{minipage}[b]{0.23\textwidth}
      \centering
      \includegraphics[scale=0.105]{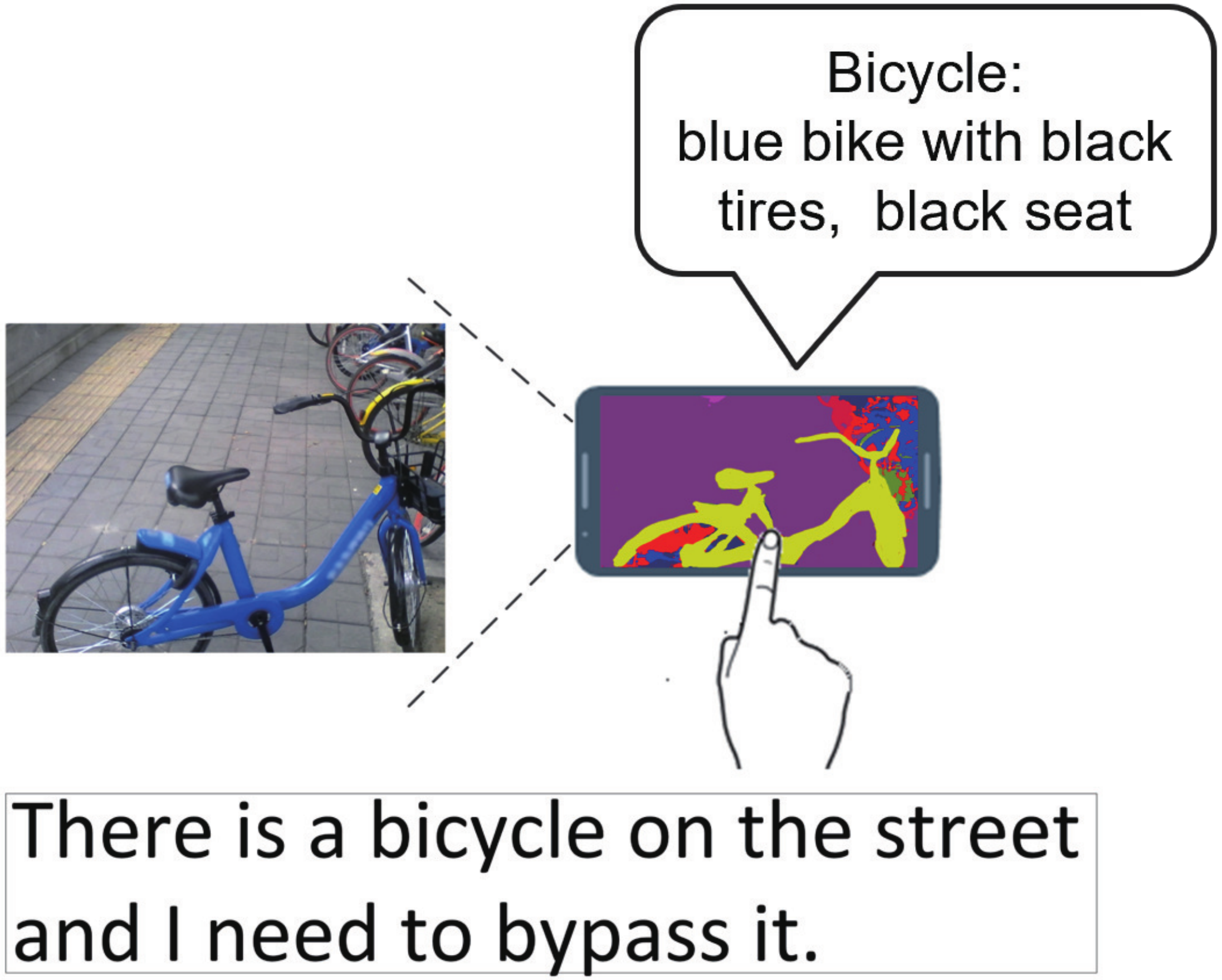}
    \end{minipage}}
  \subfigure[]{
    \centering
    \label{fig:userstudycase:b}
    \begin{minipage}[b]{0.23\textwidth}
      \centering
      \includegraphics[scale=0.105]{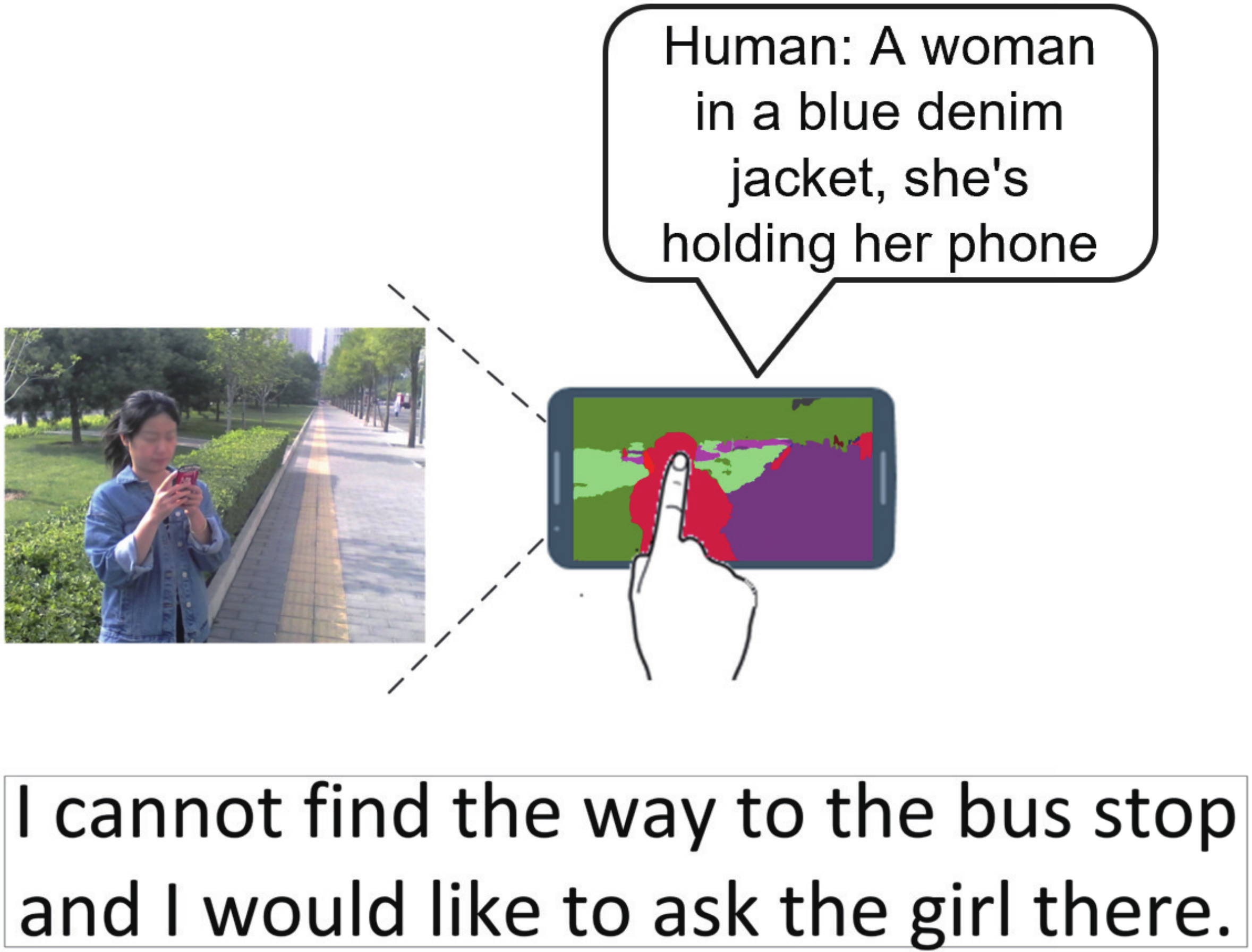}
    \end{minipage}}
  \subfigure[]{
    \centering
    \label{fig:userstudycase:d}
    \begin{minipage}[b]{0.23\textwidth}
      \centering
      \includegraphics[scale=0.105]{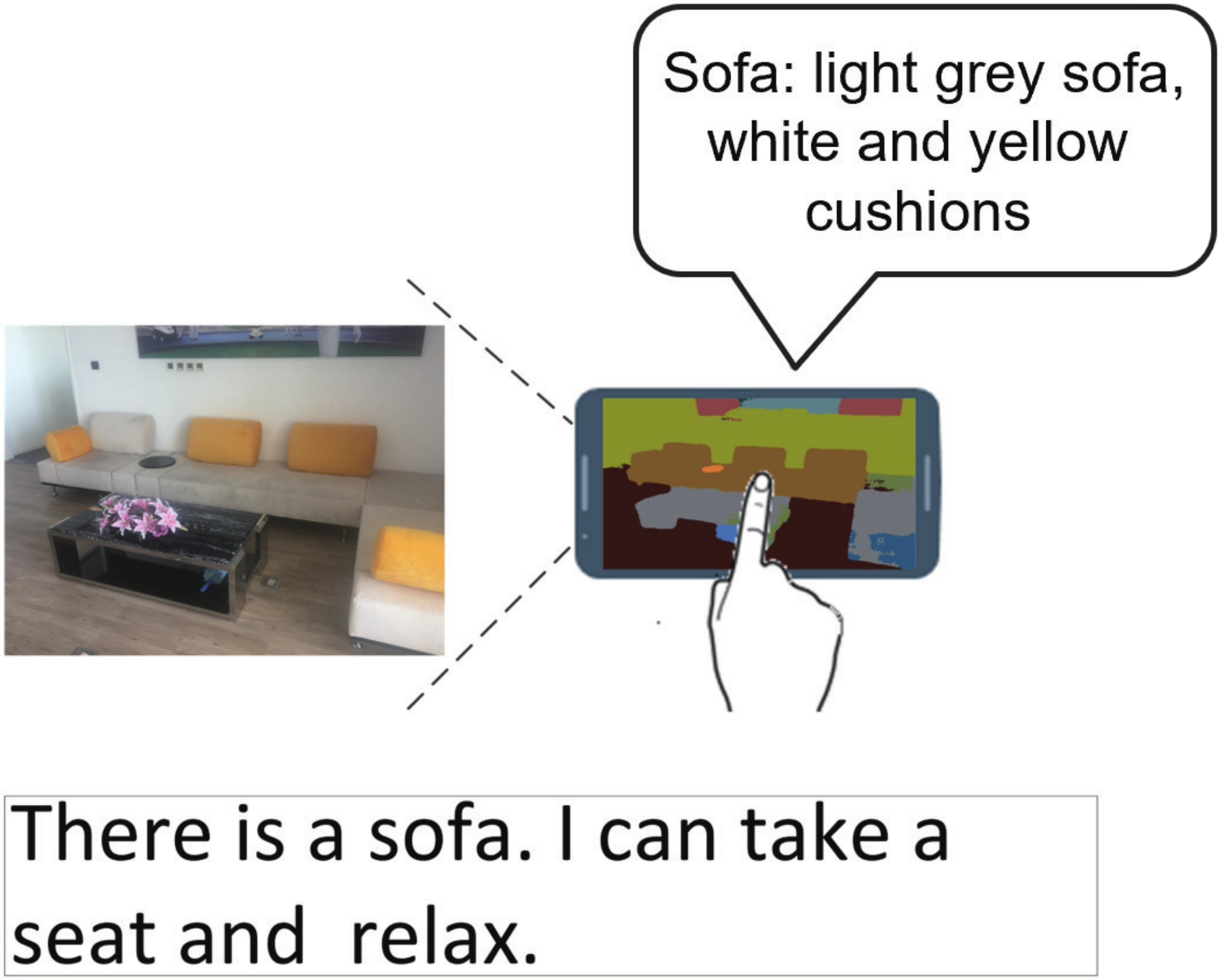}
    \end{minipage}}
  \subfigure[]{
    \centering
    \label{fig:userstudycase:e}
    \begin{minipage}[b]{0.23\textwidth}
      \centering
      \includegraphics[scale=0.105]{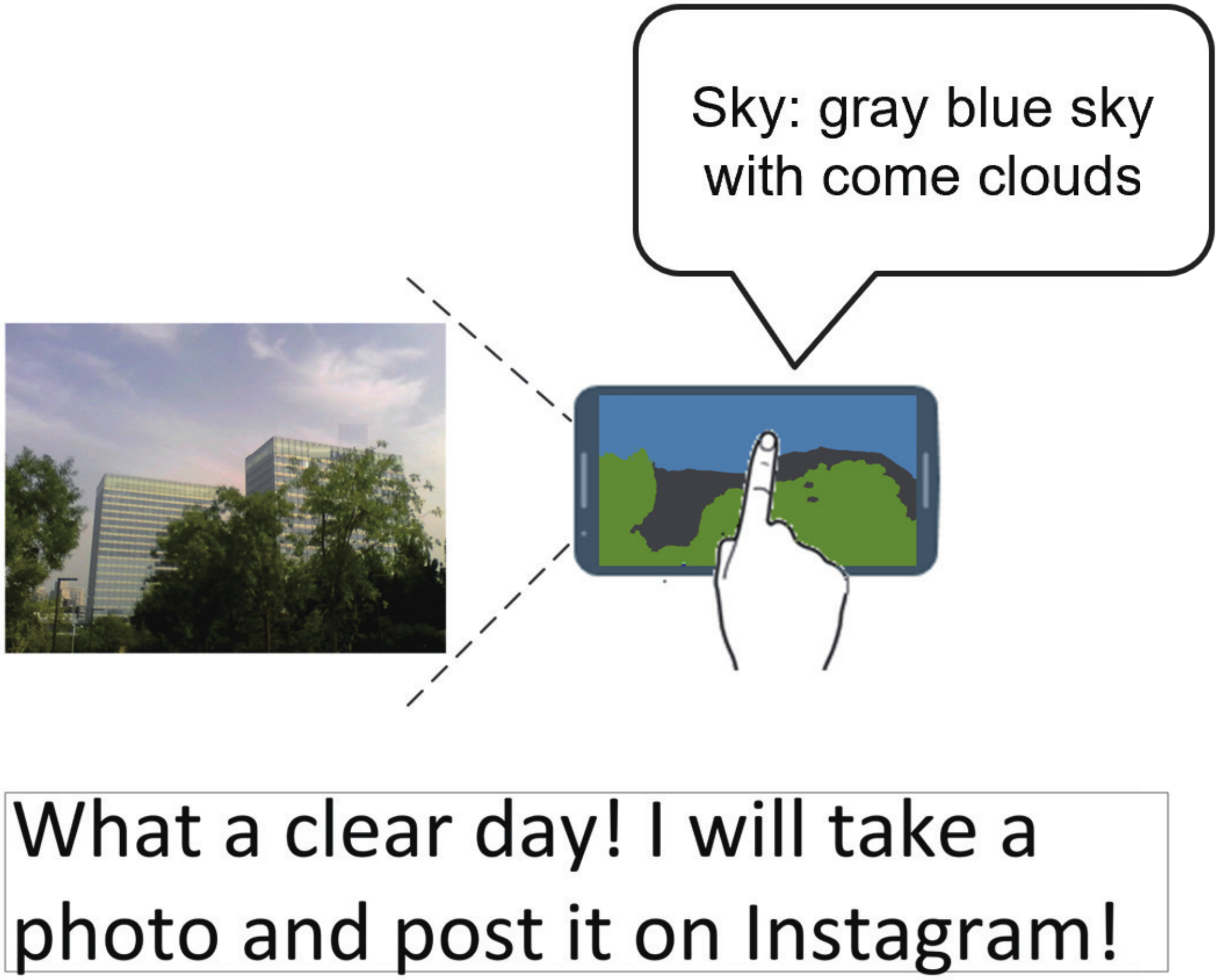}
    \end{minipage}}
  \caption{. Results of the users performing specific tasks with our system in different scenes. The users were performing (a) obstacle avoidance, (b) way asking, (c) seat finding and (d) photo taking respectively. The analyzed results and the users' feedback are listed.}
  \label{fig:userstudycase}
\end{figure}

During the evaluation, we also asked the participants to perform some tasks designed in the four scenes, such as obstacle avoidance, way asking, seat finding and even photo taking. 
Some recorded results are shown in Fig.~\ref{fig:userstudycase}. 
From the feedback of the users, it can be seen that our system is helpful in not only 
general environmental understanding but also in performing many specific daily tasks.

Generally speaking, our system works effectively 
and indeed helps the visually impaired participants 
to perceive the surrounding environment during the evaluation. 

\section{CONCLUSIONS}\label{sec:con}

This paper develops a system to help visually impaired people to sense the surrounding environment. 
This paper carefully designed the hardware for capturing the scene and giving feedback, 
as well as the retrieval of the analysis results and the supporting algorithm, 
such that the users can easily get a global description of the scene in front of them by long press action, 
perceive each object's category in the scene just by tap and swipe action, 
and acquire further knowledge of a certain object by double tap action. 
The LVLM affords the system cross-scene understanding capability, 
the segmentation model provides external knowledge to mitigate hallucinations in the LVLM 
to help visually impaired people more accurately perceive the world.
We conduct experiments and user studies in different scenes, 
and results show that our system is able to provide visually impaired people with abundant and detailed information as they expected. 
The ease-of-use features of our system make it a useful assistant for people with vision impairment. 
Besides perceiving the surroundings, it can also help them carry out various daily tasks, 
which greatly enhances the quality of their lives.

\bibliographystyle{IEEEtran}
\bibliography{IEEEabrv, IEEEexample}

\addtolength{\textheight}{-12cm}   




\end{CJK}
\end{document}